\documentclass[10pt,twocolumn,letterpaper]{article}

\usepackage[pagenumbers]{cvpr} % To force page numbers, e.g. for an arXiv version

\usepackage{graphicx}
\usepackage{amsmath}
\usepackage{amssymb}
\usepackage{booktabs}

\usepackage[pagebackref,breaklinks,colorlinks]{hyperref}

\usepackage[capitalize]{cleveref}
\crefname{section}{Sec.}{Secs.}
\Crefname{section}{Section}{Sections}
\Crefname{table}{Table}{Tables}
\crefname{table}{Tab.}{Tabs.}

\def\cvprPaperID{29} % *** Enter the CVPR Paper ID here
\def\confName{CVPR}
\def\confYear{2022}

\begin{document}

%%%%%%%%% TITLE - PLEASE UPDATE
\title{Spatial-temporal Transformer for Affective Behavior Analysis}

\author{Peng Zou\\
Institute of Artificial Intelligence\\
Hefei Comprehensive National Science Center\\
Hefei, China \\
{\tt\small zonepg@mail.ustc.edu.cn}
\and
Rui Wang\\
Institute of Artificial Intelligence\\
Hefei Comprehensive National Science Center\\
Hefei, China \\
{\tt\small qq123rw@gmail.com}
\and
Kehua Wen\\
Institute of Artificial Intelligence\\
Hefei Comprehensive National Science Center\\
Hefei, China \\
{\tt\small 494198544@qq.com}
\and
Yasi Peng\\
Institute of Artificial Intelligence\\
Hefei Comprehensive National Science Center\\
Hefei, China \\
{\tt\small 1558778695@qq.com}
% For a paper whose authors are all at the same institution,
% omit the following lines up until the closing ``}''.
% Additional authors and addresses can be added with ``\and'',
% just like the second author.
% To save space, use either the email address or home page, not both
\and
Xiao Sun\thanks{Corresponding author}\\
Institute of Artificial Intelligence\\
Hefei Comprehensive National Science Center\\
Hefei, China \\
{\tt\small sunx@iai.ustc.edu.cn}
}
\maketitle

%%%%%%%%% ABSTRACT
\begin{abstract}
The in-the-wild affective behavior analysis has been an important study. In this paper, we submit our solutions for the 5th Workshop and Competition on Affective Behavior Analysis in-the-wild (ABAW), which includes V-A Estimation, Facial Expression Classification and AU Detection Sub-challenges. We propose a Transformer Encoder with Multi-Head Attention framework to learn the distribution of both the spatial and temporal features. Besides, there are virious effective data augmentation strategies employed to alleviate the problems of sample imbalance during model training. The results fully demonstrate the effectiveness of our proposed model based on the Aff-Wild2 dataset.
\end{abstract}

%%%%%%%%% BODY TEXT
\section{INTRODUCTION}
\label{sec:intro}
The in-the-wild affective behavior analysis is a study to understand individuals' emotions and moods through their facial expressions, action behavior and physical characteristics. It has been an important research direction in the fields of mental health treatment, human-computer interaction, and marketing research\cite{ferrer2020role,baden2019impact,nguyen2023affective,kowalczuk2021cognitive}. Usually, three main representations, i.e., Action Units (AU), Valence-Arousal (VA), and the basic facial expressions(e.g., happy, sad, and neutral) are used to gain insight into how individuals express and experience emotions. The Facial Action Coding System (FACS) proposed by Ekman and Friesen\cite{ekman1978facial} has been widely applied to recognize specific emotions based on AU. Valence and Arousal are two dimensions used to describe and measure emotions. Different from the previous datasets\cite{lucey2010extended,mollahosseini2017affectnet,zhang2016multimodal} with only one or two of the three representations, D. Kollias et. al. \cite{kollias2023abaw,kollias2022abaw,kollias2022abaw1,kollias2021distribution,kollias2021analysing,kollias2021affect,kollias2020analysing,kollias2019expression,kollias2019face,kollias2019deep,zafeiriou2017aff} proposed Aff-Wild2 containing the above three representations in the wild. There are various challenges in this dataset, such as head poses, ages, sex, etc. The $ 5_{th} $ ABAW Competition 2023 consists of four tracks: valence-arousal (V-A) estimation, expression classification, action unit (AU) detection, emotional reaction intensity (ERI) estimation. The main goal of this competiiton is to understand people's feelings and develop a better platform for HCI systems.

In this paper, we propose our method to address the first three challenges, i.e., V-A estimation, expression classification and AU detection. To provide a solution to V-A estimation, Meng et.al \cite{meng2022multi} utilized the multi-modal fusion and proposed a temporal encoder and a model ensemble strategy to make their method perform better. Zhang et.al \cite{zhang2022transformer} utilized a unified transformer-based multimodal framework to fully extract the features of spoken words, speech prosody, and facial expression in the Aff-Wild2 dataset. And this method achieved good results for the challenges of expression classification and AU detection. In this study, we train ResNet-50\cite{he2016deep} using VGGFace2 dataset\cite{cao2018vggface2}. Different from the original facial expression dataset, the VGGFace2 is made of around 3.31 million images divided into 9131 classes, each representing a different person identity. And this dataset contains higher quality images and includes both regular human faces and precise facial expressions. We utilize this pre-trained ResNet-50 to extract visual features. Inspired by Chen et.al\cite{chen2020transformer} We also adopt the Transformer Encoder with Multi-Head Attention framework to fully extract visual features and predict the probability as shown in Fig \ref{Fig3}.

The main contributiosn of our proposed method can be summarized as:
\begin{itemize}
	\item [(1)]We have tried to fully extract new visual features to directly lead to significant accuracy improvements of both the baseline model and our inference model;
	\item [(2)]The Transformer Encoder with Multi-Head Attention framework can learn the 
	distribution of both the spatial and temporal features.
	\item [(3)]There are virious effective data augmentation strategies employed to alleviate the problems of sample imbalance during model training and prevent the model from learning biased subject characters. These strategies successfully improve the generalization accuracies on the unseen test data.
\end{itemize}

The remainder of this paper is organized as follows. Related works are introduced in Section \ref{s2}. Section \ref{s3} describes the details of our proposed method. Section \ref{s4} presents the implementation and experiments to evaluate the proposed method. And our work is concluded in Section \ref{s5}.
\section{RELATED WORKS}
\label{s2}
\subsection{V-A Estimation}
V-A estimation (Valence-Arousal estimation) is a subfield of affective computing that focuses on automatically predicting the emotional valence (i.e., positive or negative) and arousal (i.e., level of intensity) of human communication. Usually, there are four approches to address this task: supervised learning, deep learning, multimodal fusion and transfer learning. Zhang et.al \cite{zhang2022continuous} utilized a cross-modal co-attention model for V-A estimation using visual-audio-linguistic information. A two-stage strategy proposed by Nguyen\cite{nguyen2022ensemble} was introdeced. This method extracted new features and used the ensamble approach.
\subsection{Facial Expression Classification}
Facial expression classification has been a popular research topic in the field of computer vision and machine learning. There have been many studies and approaches proposed to address this problem. In the $ 3_{th} $ ABAW Competition 2022, Zhang et.al \cite{zhang2022transformer} introduced a transformer-based fusion module that extracted the visual features and the dynamic spatial-temporal features and ranked the $ 1_{st} $ place. Jeong et.al\cite{jeong2022facial} adopted an extended DAN model to address the facial expression.
\subsection{AU Detection} 
AU detection involves the identification and tracking of subtle facial movements that correspond to specific emotional states or expressions. The improved IResnet100 introduced by Jiang et.al\cite{jiang2022facial} was utilized to address the AU detection task. And Wang et.al\cite{wang2022multi} proposed a action units correlation module to learn relationships between each AU labels and proved the effectiveness of the method.
\section{PROPOSED METHOD}
\label{s3}
\begin{figure*}[htbp] %H为当前位置，!htb为忽略美学标准，htbp为浮动图形
	\centering %图片居中
	\includegraphics[width=1.0\textwidth]{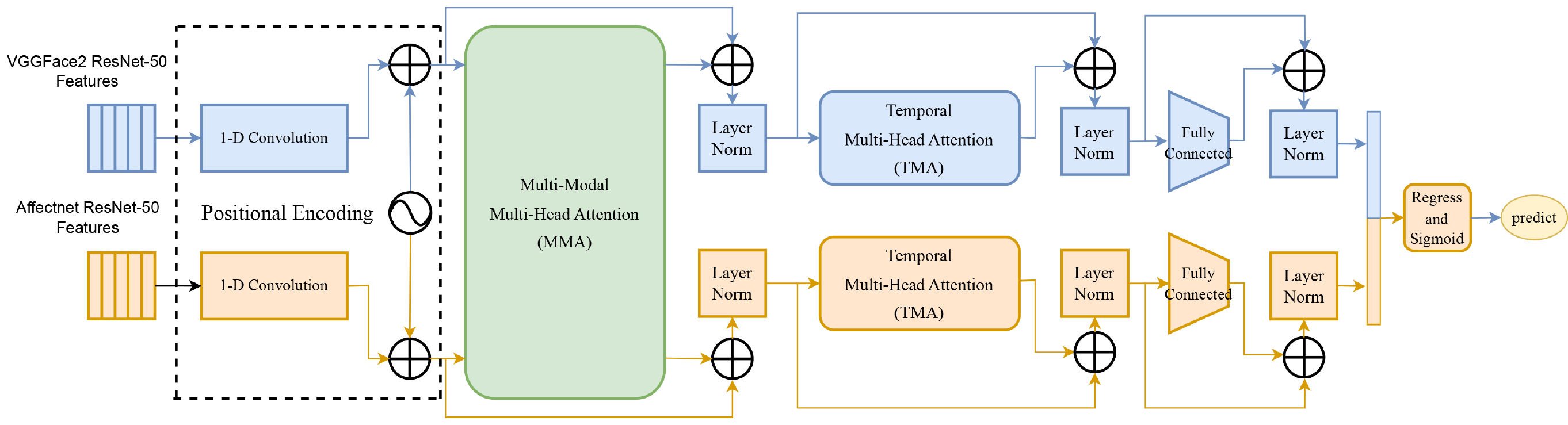} 
	\caption{The overall architecture of our proposed method. Two kinds of features are used, i.e., the VGGFace2 ResNet-50 features and the Affectnet ResNet-50 features. } %最终文档中希望显示的图片标题
	\label{Fig3} %用于文内引用的标签
\end{figure*}

The overview of our proposed method is illustrated in Fig \ref{Fig3}. The detailed process is described as follows.

\textbf{Self-attention Mechanism}. In 2016, self-attention was proposed by Vaswani et.al \cite{vaswani2017attention}. In this paper, we utilize this mechanism to fully obtain the spatial and temporal features. Given the input visual feature sequence $ X^{v}=\left\{X_{i}^{v} \in \mathbb{R}^{d_{v}}|i=1, \ldots,|T|\right\} $. Then, the output $ O^{v}=\left\{O_{i}^{v} \in \mathbb{R}^{d_{v}}|i=1, \ldots,|T|\right\} $  can be calculated as follows:
{\setlength\abovedisplayskip{0.7pt}
\setlength\belowdisplayskip{0.7pt}
\begin{equation}
	O^{v}=ATT_{v}\left(X^{v}\right)
\end{equation}}
where \textit{$ d_{v} $}  donates the dimension of the visual feature sequence and $ |T| $ is the max time step. And the detailed calculation is given by Eqn \ref{q1}:
{\setlength\abovedisplayskip{0.7pt}
	\setlength\belowdisplayskip{0.7pt}
	\begin{equation}
	\label{q1}
	O^{v}=softmax(\frac{Q^{v}\cdot K^{vT}}{\sqrt{d_{k}}})\cdot V^{v}
	\end{equation}}
The $ Q^{v}, K^{v}, V^{v} $ represent the queries, values and keys matrix mapped by the input acoustic sequence \textit{$ X_{a} $}, respectively.

\textbf{Visual Feature Embedding}. We use a 1-D temporal convolution network to help our model better capture temporal information for each feature vector. For the outputs of the embedded sequence, we add the positional encoding. The length of the final output embedded
vector is $ d_{m} $.

\textbf{Temporal Multi-head Attention (TMA)}. We utilize this method to capture the temporal dependency. Suppose the timestep is $ t\in[1,T]  $, it is calculated as follows:
{\setlength\abovedisplayskip{0.7pt}
	\setlength\belowdisplayskip{0.7pt}
	\begin{equation}
	\begin{aligned}
	Q_{v} &= Concat(Q^{1},Q^{T},...,Q^{T})W^{Q}\\
	K_{v} &= Concat(K^{1},K^{T},...,K^{T})W^{K}\\
	V_{v} &= Concat(V^{1},V^{T},...,V^{T})W^{V}\\
	\end{aligned}
	\end{equation}}

%-------------------------------------------------------------------------
\section{EXPERIMENTS}
\label{s4}
\subsection{Dataset}
In our study, we conduct experiments on the large-scale in-the-wild Aff-Wild2 database. And the given images are cropped and aligned with an input size of 112$ \times $112 in RGB from each video clip. For V-A estimation, the dataset consists of 594 videos of around 3M frames of 584 subjects annotated in terms of valence and arousal. Data is used containing annotations on 6 basic expressions including Anger, Disgust, Fear, Happiness, Sadness, Surprise, plus Neutral state, and Other which denotes emotional expressions other than the 6 basic states in the expression classficaion task. And for AU estimation, the Aff-Wild2 database is audiovisual (A/V) and in total consists of 547 videos of around 2.7M frames that are annotated in terms of 12 action units.
\subsection{Training Details}
Our proposed network is trained on a GeForce RTX 3090 GPU based on the opensource PyTorch platform. We use Adam optimization \cite{kingma2014adam} to update the weights.  In our experiments, the facial images all have a size of 112$  \times $112. The batch size of 32 is trained during the training process and the batch size of 16 is used for test. Our model can automatically learn 30 epochs and save the best performance on the validation set.

\begin{table}[]	
	\label{qq}
	\caption{The results of three sub-challenges.}
	\centering
	\begin{tabular}{cc}
		\hline
		Sub-Challenge       & Metric(\%) \\ \hline
		V-A Estimation      & 54.5      \\ 
		EXPR Classification & 30.4      \\ 
		AU Detection        & 38.9      \\ \hline
	\end{tabular}
\end{table}

\subsection{Results}
We report results by CCC and $F_1$ score for the three sub-challenges in table 1 on the validation set. For the facial expression classfication, the authors for the baseline\cite{kollias2022abaw} perform the pre-trained VGG16 network on the VGGFACE dataset and get softmax probabilities for the 8 expression predictions. In our proposed model, there are virious effective data augmentation strategies employed to alleviate the problems of sample imbalance during model training. And we try to extract visual features and the Transformer Encoder with Multi-Head Attention framework is proposed to learn the 
distribution of both the spatial and temporal features. We obtain the results of 54.5\%, 30.4\%, 38.9\% for the validation sets of V-A Estimation, Facial Expression Classification and AU Detection Sub-challenges, respectively. The results fully demonstrate the effectiveness of our proposed model.

\section{CONCLUSION}
\label{s5}
In this paper, we submit our solutions for the 5th Workshop and Competition on Affective Behavior Analysis in-the-wild (ABAW). We have tried to extract new visual features and the Transformer Encoder with Multi-Head Attention framework is proposed to learn the 
distribution of both the spatial and temporal features. Besides, there are virious effective data augmentation strategies employed to alleviate the problems of sample imbalance during model training. We obtain the results of 54.5\%, 30.4\%, 38.9\% for the validation sets of V-A Estimation, Facial Expression Classification and AU Detection Sub-challenges, respectively.

%%%%%%%% REFERENCES
{\small
\bibliographystyle{ieee_fullname}
\bibliography{egbib}

}

\end{document}